# Framework for Passenger Seat Availability Using Face Detection in Passenger Bus

Khawar Islam and Uzma Afzal

*Abstract*—Advancements in Intelligent Transportation System (IES) improve passenger traveling by providing information systems for bus arrival time and counting the number of passengers and buses in cities. Passengers still face bus waiting and seat unavailability issues which have adverse effects on traffic management and controlling authority. We propose a Face Detection based Framework (FDF) to determine passenger seat availability in a camera-equipped bus through face detection which is based on background subtraction to count empty, filled, and total seats. FDF has an integrated smartphone Passenger Application (PA) to identify the nearest bus stop. We evaluate FDF in a live test environment and results show that it gives 90% accuracy. We believe our results have the potential to address traffic management concerns and assist passengers to save their valuable time.

| ACRONYM | DEFINITION |
|---|---|
| PSA | Passenger Seat Availability |
| PA | Passenger App |
| APC | Automatic Passenger Counting |
| ERF | Electronic Registering Fareboxes |
| IS | Infrared Sensor |
| GSM | Global System for Mobile Communication |
| BS | Barrier Sensor |
| TMS | Treadle Mat Sensor |
| WIM | Weight In Motion |
| OPS | Optimal Sensor |
| AVL | Automatic Vehicle Location |

## I. INTRODUCTION

In the development of traffic management systems, an intelligent APC system was developed to reduce a time of passenger seat availability, security issues and provide more efficient method to calculate ridership data and use optimal resources for transportation sectors. Currently, the common problems are such as safety, traffic congestion [4,5] automatic passenger counting systems for public transport, including offline systems [6] load monitoring systems for passenger counting [7] loss of public space at bus stations [8] public transport inadequacy, daily ridership data [9] high infrastructure costs [10] and etc. Especially for APC are designing and implementing embedded based system for passenger detection [11]. Contour feature is used to detect a passenger under a complicated situation. Tao et al [12] proposed a novel approach based on a clustering method for counting passengers at the entrance gate of a bus equipped with a camera. The research behind this study is to solve the current issue and suggest visionary approach in the transportation system, such as a passenger waiting for a bus, the traffic congestion problem during heavy passenger loads on the bus station, provide a high security at lower cost and cheap maintenance. The study provides a new mechanism of APC and PSA. The information of nearby buses are update in PA to book a seat earlier using a supported device such as smartphones. Furthermore, the booking framework uses the identification number of people to maintain history. However, the current mechanism of APC doesn't provide an excellent solution for finding bus station and maintain a seat availability for passenger, does handle passenger load on peak time, and does not provide any economic benefit. The proposed framework is less cost and highest performance with the comparison of existing systems [11] [12].

To solve the above-mentioned issues and achieve the significant enlargement in machine learning, the algorithm has created as well as face detection application has developed to overcome this problem. The present study proposes and develops a new framework based on face detection. Our framework detects each face of a passenger in a bus, and the data of filled and empty seats will be sent to the data center. The data center calculates empty seats, and this information is frequently updated in PA through the internet. PA is based on innovative technology and sees a live view of buses on a map. Furthermore, in the proposed framework, the passenger looks the empty seats of upcoming buses. This research also provides a cheaper solution to count ridership on a daily basis, high security and low-cost maintenance.

The literature review (presented in next section) discovers that no one used the facility of the camera which is already equipped in the bus to count the number of passengers, although face detection is one of the most explored machine learning technique in recent years [42]. Face detection also refers to face-priority (autofocus), it is the function of the camera which detects human faces. Majority of machine learning techniques are used to achieve face recognition feature which has many successful applications to the different problem domain. Organizations use face detection for a security purpose, banks use a biometric approach for money transaction [39]. In recent studies, smartphones are also camera enabled which recognize a human face for payment and phone unlock. In [40], the authors propose a Viola-Jones algorithm to detect faces which further used for feature extraction. The perspective of Intelligent Vehicles. In [41], the authors implement a passenger management system based on face detection using the AdaBoost method. In this context, we posed and addressed the following research questions in this paper;

**RQ1:** What is the possible value which face detection method can bring to APC?

**RQ2:** What are the advantages of the proposed solution over existing solutions?

To answer the RQ1, we FDF improves existing PT system. It reduces the passenger waiting time for a bus, helps to counter the traffic congestion propose a novel Face Detection based Framework (FDF) which detects passenger faces in a bus. An algorithm is used to recognize the occupancy of the seats (filled or empty). A number of empty seats in a bus are also calculated. FDF implements a user interface through a mobile-based Passenger App (PA). We evaluated FDF through simulation and implementation. The results of the simulation advocate the use of face detection to improve APC. To answer RQ2, we compare the results of FDF and existing APC solutions. The comparison results prove that problem in heavy passenger loads on the bus station, moreover its maintenance is cheap. We focused on passenger's sit on the bus seat. Figure 1 illustrates a side of bus and camera focus on a bus seat. Located in an internal area and front of the bus which captures images the passenger sited in the bus and responds PSA.

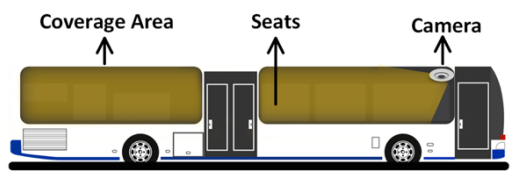

Figure 1. Side view of Bus using the normal lighting camera

*A. CONTRIBUTION*

With the intention of achieving, the issues of transportation mentioned above and encouraged with some suited work [11], [12]. We purpose new APC and seat availability based on face detection and seat availability algorithm. First, our algorithm supports a mechanism to search the nearest bus, which will arrive at a bus station. Second, we adopt a mechanism for suggesting the passenger to another bus station if the current bus is full of seats. We proposed a framework of a passenger sitting in a bus seat such that each bus is connected with the framework. Each bus obtains information from the PA. As a result, smooth traveling of passengers at low cost and reduce traffic congestion, bus-waiting issues. Our framework achieves better in the term of performance and the time saving with other systems. We evaluated the framework performance through simulation and implementation. The results of the simulation are close to our result and better performed than the other systems. The proposed framework reduces the passenger's time to find or wait for a bus and minimize the costly maintenance, solve security and traffic congestion problems.

The primary contributions of our research work as follows:
- We present the first face detection based solution to improve the existing APC.
- We implement the proposed framework in a live testing environment.
- We compare the proposed FDF with existing APC solutions to prove its usability.

The rest of the paper is organized as follows. Section 2 presents the relevant background, Section 3 describes our proposed framework FDF. While the experimental setup and data collection procedure are in Section 4. Section 5 discusses the results and evaluation. We conclude our work in Section 6.

## II. RELATED AND BACKGROUND WORK

In this section, we present the relevant background related to the automation of public transport systems. We also reviewed the applications of face detection to the different domain. Our study reveals that real-time vehicle prediction is the most explored domain [1,3,4,6,7,8] along with the bus ticketing system [5], passenger detection system [10, 11], vehicle tracking system [12, 19] and sensor data system [2].

*A. VEHICLE DETECTION SYSTEMS*

In this section, we discuss the automation of transport systems which are related to detection of the vehicles. Attouche et al. [13] propose a prediction based system to evaluate the collision and unhooking incidents at the highway. The aim of the study is to measure the number of trucks running on the highway. The system immediately generates an alarm to the truck when the number of trucks exceeds the predefined threshold value. The proposed solution solves the congestion problem and helps to reduce the number of accidents. Weiming et al [14] present a probability model based on the fuzzy and neural network which learns different patterns of the accidents to predict the road accidents. The model uses sample data of 3-D vehicle tracking trajectories. Sayanan et al. [15] present a literature review of vehicle detection and tracking systems under different circumstances. This review is based on stereo vision and monocular configurations in a 3-D domain including motion and filtering. Shiva et al. [16] propose an algorithm to count and verify the vehicles in a video frame. They recommend texture based features to improve the counting of the vehicle. The proposed method shows good accuracy in a real environment such as highways. Afshin et al. [17] estimate traffic flow and predict the short-term traffic flows for transportation. Monte Carlo simulation is used for uncertainties. The result shows the accurate prediction of traffic flows up to 30 min under normal condition.

*B. VEHICLE ARRIVAL SYSTEMS*

In this section, we discuss the automation of transport systems which are related to different parameters of vehicle arrival such as arrival time. Fangzhou et al. [18] present a decision support system based on clustering which to give real-time predictions. Historical data of different routes are used to train the predictive model which predicts the time of vehicles arrival and marks the anomalous operations. Results show two hours gap in overhead performance because of the challenging nature of the problem, i.e., prediction of vehicle arrival time. Historical data which was collected in transit hub system includes both static and dynamic feeds. System is deployed on private cloud. Bratislav et al. [19] propose a sensor less Automatic Vehicle Location (AVL) Results show that minimal configuration of networks and sensor provide sufficient data for consistent analysis of model prediction which ultimately generates accurate prediction of bus arrival. Ranhee and Laurence [20] also predict arrival time of vehicles based on the Artificial Neural Network (ANN) and regression. Data from Houston and taxes bus routes is used to train and test the model. The ANN outperforms in terms of historical data where regression model gives accuracy in terms of prediction. James et al. [21] develop an automatic transit system which is integrated with smartphone based system called easy tracker. GPS (Global

Positing System) helps to trace bus routes served, bus stops and bus schedule. They used an algorithm to predict arrival time for upcoming stops. Existing dataset of routes, arrival time, stops time is used to train the model and predict arrival time of the bus. Sudhakar and Rashmi [22] also use GPS to locate bus coordinates and send information to passenger about bus. GSM sends data to central monitoring server to calculate bus arrival time and send relevant information to passengers. Padmanaban et al. [23] develop a model based on algorithms for bus arrival time, they also include delay time to improve the accuracy of the results. Kalman technique is used to predict bus arrival time which improves the overall performance of the proposed system. [24] Introduces micro controller design and propose a smart transportation system. Passenger uses the system by entering current location of the nearest bus stop when traveling in a bus which saves passenger time to wait hours for a bus at the bus station.

Swati Chandurkar et al. [25] present a passenger information system based on bus monitoring concepts. This information system works as a standalone application to provide an interface to the passengers who can see the covered route of the GPS equipped. GPS is used to send information to the centralized control system. Johar Amita et al. [26] propose an ANN based approach to predict bus travel time. ANN model is trained using the data of two urban routes. Results are evaluated on the basis of correlation coefficient, standard deviation and statistical test. ANN gives better than regression in terms of performance. Stuart et al. [27] propose a WAP enabled system specially for mobile users to receive bus arrival information. Shital et al. [28] also present hardware mechanism interaction with GPS and GSM services. System is connected with a car alarm and user gets alerts on his smartphone. In [29], authors implement a low-cost application for a safe automobile. Jain et al. [30] present multiple approaches for bus arrival time, i.e., K-Nearest, ANN. Historical data of real time routes of bus arrival is collected to learn the model. Prediction accuracy of both algorithms is less than 12%. Santa et al. [31] present a method to predict bus arrival time for developing countries where data collection is a big challenge. They proposed a model which captures vehicle time and road condition based on historical data. SVM and ANN performed very well for time prediction.

## C. PEDESTRIAN DETECTION AND FLOW

Wilson and Svetha [32] present pedestrian detection system through different scenes captures by mobile and surveillance bus. Mate Szarvas et al [33] uses convolutional neural network for detection of pedestrian. Junfeng Ge et al [34] present a pedestrian detection system based on monocular vision system for night. Fanping Bu et al [35] use sensing technology to detect pedestrian in transmitting bus. Anelia et al [36] use deep network cascades approach to detect real-time pedestrian.

## D. PASSENGER DETECTION AND INTEGRATION SYSTEM

Designing and implementation of an embedded system for passenger detection is an emerging track for researchers [37]. Contour feature is used to detect a passenger under the complicated situation. Tao et al [12] propose a clustering-based method for counting passenger in a bus with a single camera. After an in-depth study of different research papers on passenger detection. Table I shows the summary of existing systems and efficiency of systems.

Table I intends to give a concise summary, which compares the PSA framework against related solution and systems. We can see that PSA covers the whole process of faces detection, counts passengers faces, relevant information send to PA. This framework directly reduces the cost and maintenance and provide a proper solution traffic congestion problem, security risks, and traffic control management.

### III. FDF: A Proposed Solution

As we already mentioned, FDF (Face Detection based Framework) is based on face detection technique to improve the existing APC systems. It is divided into modules to reduce the overall complexity of APC (Figure 2). The architecture and functions of these modules are as follows:

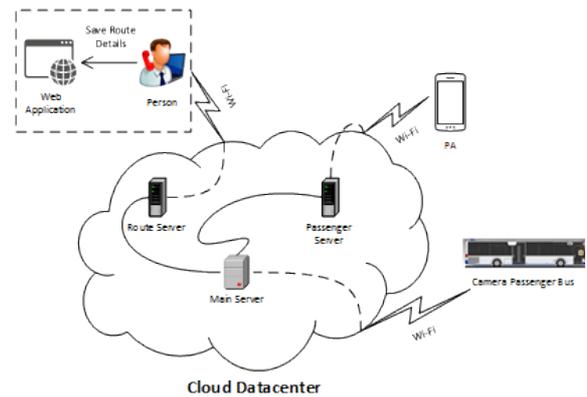

Figure 2. The architecture of the proposed Framework

- Technical Person Module (TPM): It has an interface for a technical person to register routes data according to a bus route which is further saved to a data repository (Name). Passengers used to route data to identify the current position of the bus, routes of the bus and nearest route of an upcoming bus.

TABLE I
COMPARISON OF EXISTING SYSTEMS

| Reference | Paradigm | Passenger Detection | | | | Description |
|---|---|---|---|---|---|---|
| | | Actual | Detects | Missed | Percentage (%) | |
| [25] | HOG Algorithm | 113 | 109 | 4 | 96 | Design an algorithm and software for passenger flow detection with real-time embedded equipment. |
| | | 114 | 104 | 10 | 91 | |
| [26] | Hough Transform & Fuzzy | 123 | 110 | 13 | 88 | Applied image processing techniques for real-time detection of passengers. The produced technique applied in DSP based embedded system. |
| | | 27 | 23 | 4 | 82 | |
| | | 50 | 47 | 3 | 93 | |
| | | 33 | 31 | 2 | 93 | |
| | | 233 | 211 | 22 | 89 | |

- Web Application Module (WAM): It receives route data from TPM which is connected to a cloud datacenter and update the route server. It has an interface for administrations to check the buses route data and more relevant information about the buses.
- Data Repositories: FDF has data servers to store the data. Data is distributed among them to balance the load.
  - Route Server (RS): It is specially designed to maintain the routes data. It is connected to the main server and manages all over the bus routes in cities. RS synchronizes with the main server to keep the information up-to-date and enables WAM to retrieve updated data and statistics about bus routes.
  - Passengers Server (PS): It stores and manages the images which are captured by camera-equipped bus. PS passes this data to PSA which detects the empty and fill seats.
  - Bus Station Server (BSS): It stores information about buses running in each city. It updates the movement of the bus.
  - Cloud Datacenter (CD): It plays an important role to connect different components of FDF through a wireless network. CD is a head entity of all server like RS and PS. It has a connection with a mobile app to share the bus information. The primary objective of CD is to manage other data servers and performs the quickest operations to facilitate passengers.
- Passenger Application (PA): Passenger connects with FDF through a mobile-based PA. After logging into PA, the user can see a live view of the bus, the nearest bus station details and bus availability. The passenger can select the source to a destination location in a PA, which recommends the nearest station and upcoming buses. Upon selecting the nearest bus station, PA displays the nearest buses only. PA suggests the best route to pick, timings and seat availabilities. It also shows the current position of passenger (smartphone). If a passenger books a seat, the total number of available seats also update accordingly. PA uses "CoreLocation" (Apple library) to retrieve passenger location along with its latitude, location, city, country, postal address etc.

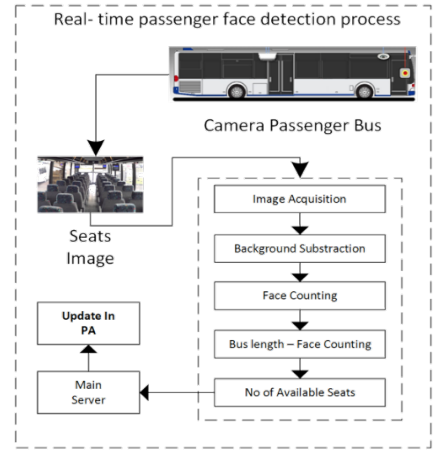

Figure 3. The process of face detection in a passenger bus

FDF assumes that passenger buses are already equipped with a WIFI enabled (3G/4G) internal camera to capture passenger images. These images stores in PS. We used MOG [34] based approach for face detection and background subtraction. Our main objective is to count the number of empty seats in the bus which is used to count the number of passengers traveling on the bus. For this, we define a line and specific part of the camera which is already mounted on the bus ceiling to detect passenger's faces when they sit on a seat. For each passenger coming to a seat, the camera takes pictures with a period of sixty seconds. The camera captures the passenger's image sited on a bus seat and tracking list of the passengers until passenger leaves the bus. The appropriate servers are also updated. PA follows a sequence of steps (Figure 3) to perform this task, i.e., detection of a passenger's face. After background subtraction, the approximate number of faces are detected which are subtracted from the total seats of the bus. A number of the available seats are updated to the main server which is connected with PA. FDF also updates in case bus is stopped.

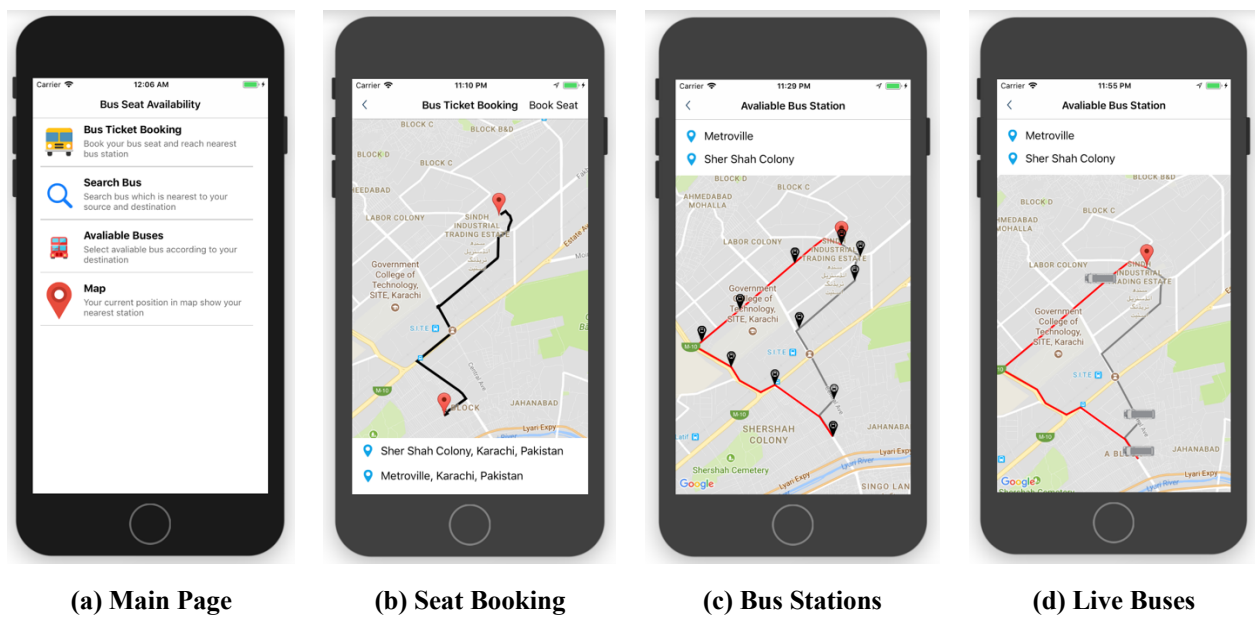

(a) Main Page    (b) Seat Booking    (c) Bus Stations    (d) Live Buses

Figure 5. A screenshot of Passenger Seat Availability application

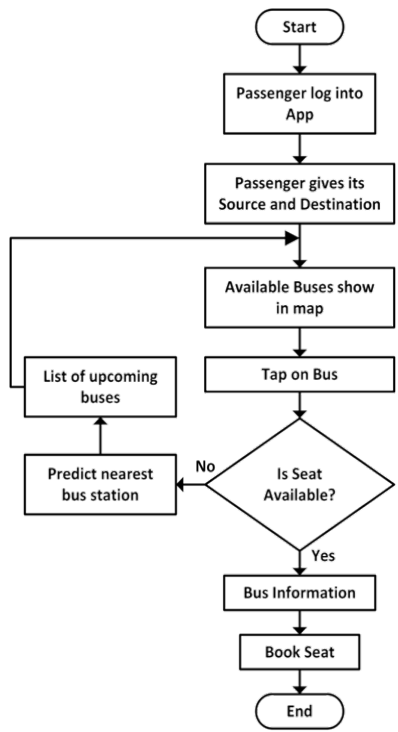

Figure 4. The flowchart of the PA operation.

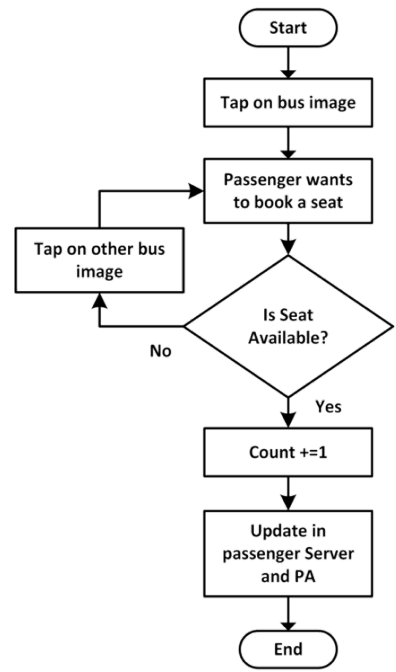

Figure 6. The proposed flowchart for update PA and passenger server.

Figure 4 shows a basic workflow of user interaction with FDF and Figure 5 presents the screenshots of PA. User interactivity starts with a login into PA using a smartphone. The passenger is asked to accept the privacy policy which grants an access to the passenger location. A background service is started when the passenger accepts the privacy policy. After a successful login, the current location of the user with respect to the latitude and longitude is updated in PS (Passenger Server). On the main page, PA prompts for options such as bus ticket booking, search bus, available bus, and map (Figure 5(a)). Each option displays a short explanation below the headings. A passenger enters the source location, i.e., the starting point of the traveling, and destination location, i.e., endpoint. After entering these locations, a map displays on the PA screen which shows the available buses (Figure 5(b)). PA draws a route between the source and destination (Figure 5(d)). Now, the passenger can click on the second option and see bus stations within the route. Figure 5(c) shows a black pin on the map to present a bus station.

The passenger can click an available bus to see the seat availability and its arrival time. If the upcoming bus is full of zero vacant seats then PA can suggest the nearest bus stations for travel. If passenger agrees to use the nearest bus station then he can again tap on the bus to see the details and reserve the vacant empty bus seat. If a seat is available for the first time then he simply sees the information of upcoming bus including empty and by bus seats, total seats, and etc books a seat.

If seats are available then Passenger can simply book the seat if seats are available, after the seat booking, a number of seats are directly updated in PS through PA. Figure 6 shows the working of FDF in case of seats unavailable. The passenger can check other buses and this process can repeat based on passenger operation.

If the seat is not available then passenger can search the nearest bus station using PA and moves to another bus stop to reserve the bus seat (Figure 7).

IV. DATA COLLECTION AND EXPERIMENTAL SETUP

In this section, we present the data collection procedure along with the experimental setup. We used different types of passenger images. These images were captured from different angles of the bus camera. The bus routes we selected to implement FDF have covered agency and management office areas of the Karachi city. These areas are controlled by the City District Government. The routes were chosen because of their diverse dynamics of traffic load, i.e., normal and peak load. We collected data over the time period Jan 2016 to Jan 2017 (7 am to 10 pm, a high number of passengers wait for a bus). Figure 7(a) (rearrange the sequence of images) shows starting point where passenger starts traveling and destination point where passenger leaves the bus. We collected data from two buses of the specified routes, they start from the same source although their routes are different; at the mid of the route, both buses are on the same track.

We performed experiments on randomly selected 50 images which show the internal view of passengers (Figure 8 and 9). We were not interested in taking more images because our primary focus was to detect the faces of the passengers rather than images. The images show different sitting behavior of the passengers, moreover, these images are captured in different modes, i.e., morning, noon, and afternoon, midnight, and full night with different image resolutions (see Figure 8 and Figure 9).

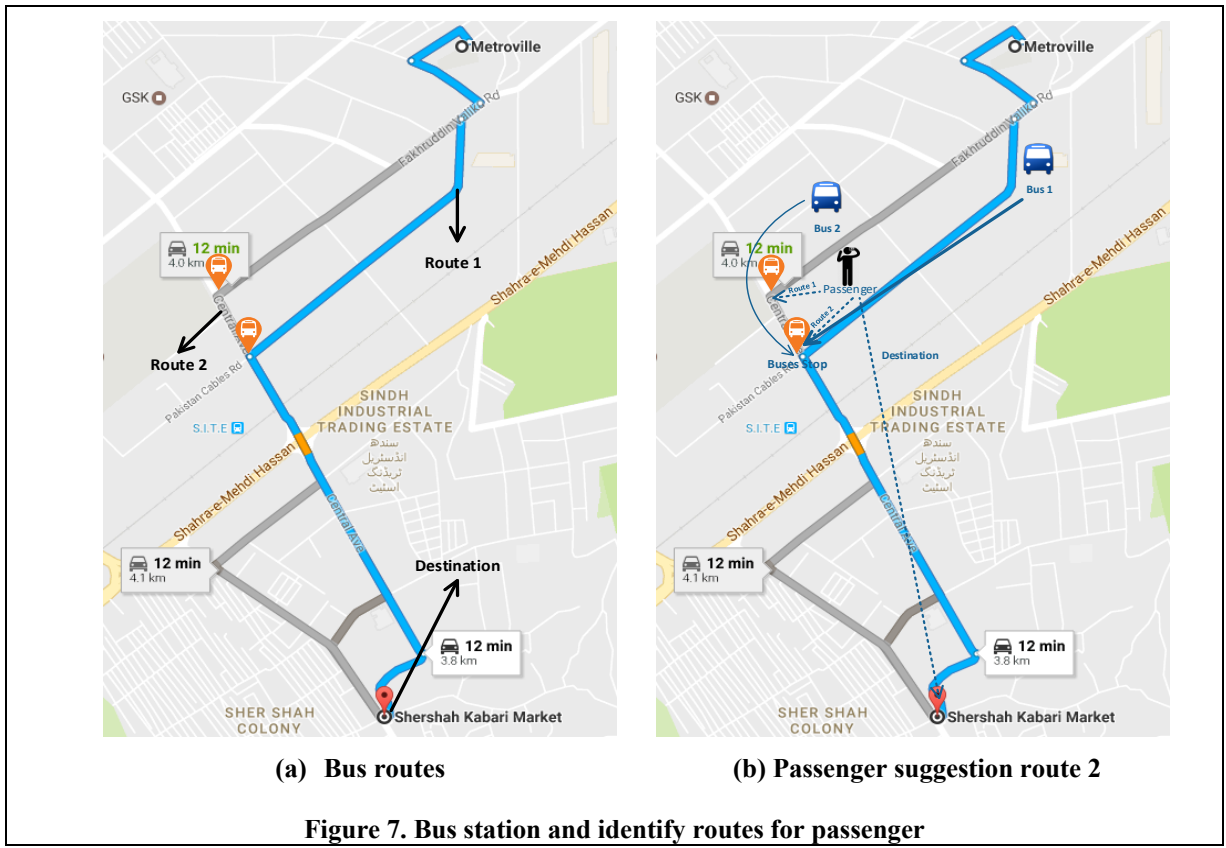

(a) Bus routes  (b) Passenger suggestion route 2

Figure 7. Bus station and identify routes for passenger

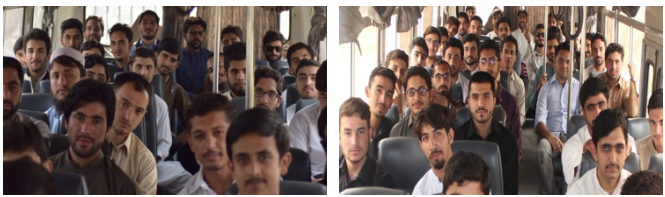
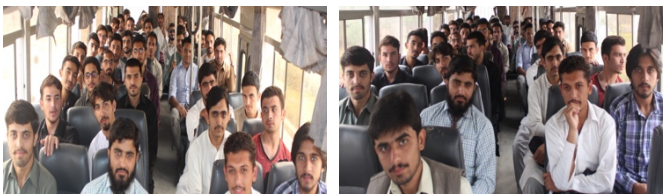

Figure 8. Example of passengers training dataset in noon and afternoon camera, which consists of images with different sizes, image resolution, and passenger behavior.

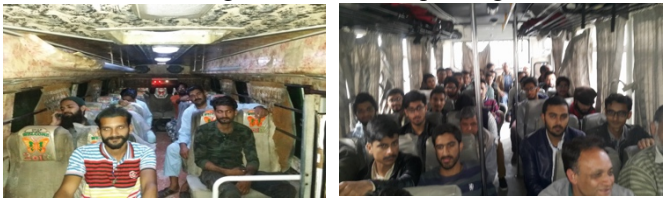
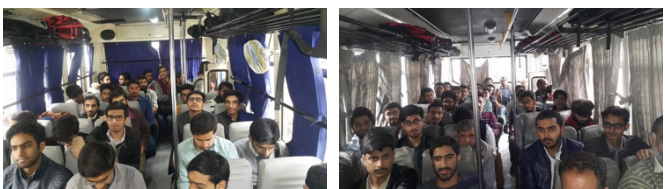

Figure 9. Example of passenger training dataset in day night camera, which consists of 50 images with different sizes, image resolution, and passenger behavior.

The images consisted of two additional parts of an internal view of the bus, i.e., lower ceiling height and high ceiling height of the bus. For both cases, a camera is mounted at a ceiling height of the bus after two seats from the head (driver seat and the left seat). We chose this particular location to mount the camera because the whole area of the internal bus can be captured from here which ultimately provides an accurate position of passengers and their faces. We ensured that every passenger face fully detects in camera using face detection. We observed that night footages of the camera were more accurate than day footages. The camera captures many pictures from different views, but we were interested in face detection and sitting behavior of the passengers. To evaluate FDF we divide the dataset into training and testing datasets.

## V. RESULT AND EVALUATION

To evaluate the framework performance, we focused the main performance factors such as a number of face detection in the context of seat availability, the cost of the servers and counting of empty seats through the framework. As we already mentioned in the experimental setup section, this research includes the normal and peak time while the bus is filled with passengers. Moreover, we took different images in day timing including morning, afternoon, midnight and night. FDF gives better results for night footages in term of performance.

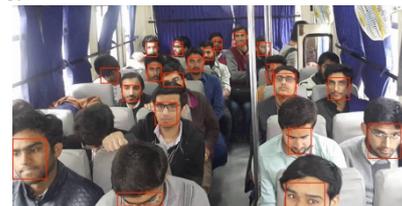

Figure 10. The internal view of a bus image in the dark afternoon

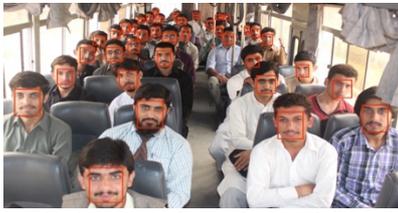

Figure 11. The internal view of a bus image in the dark morning

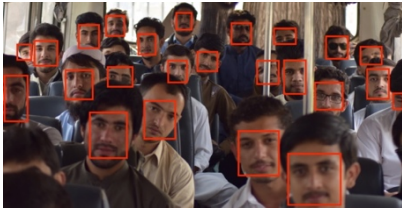

Figure 12. The internal view of a bus image in the mid-afternoon

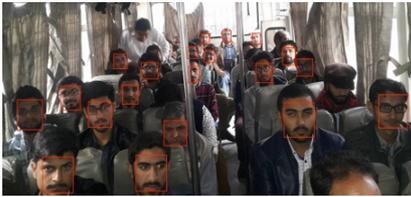

Figure 13. The internal view of a bus image in the midnight

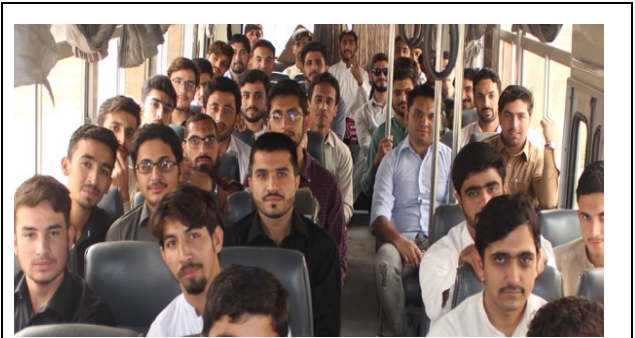

Figure 14. Color image passed in framework

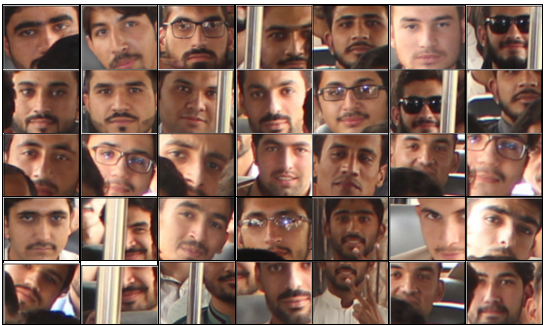

Figure 15. Faces detect from Figure 1 image

Figure 14 shows the passenger's image passed to FDF which detects and counts faces based on the eyes. Figure 15 shows the separation of the faces.

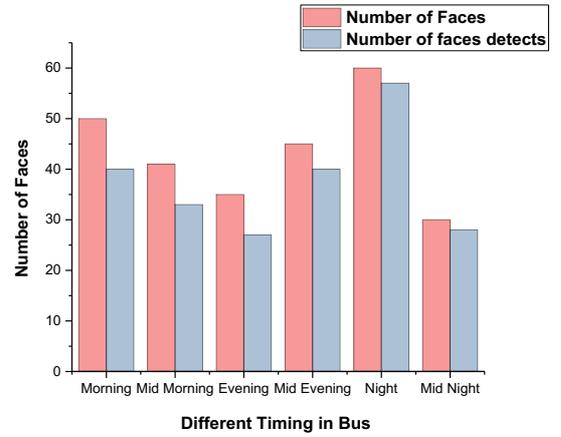

Figure 16. Evaluating the performance of the framework in different timing at day.

Figure 16 shows the comparison graph between actual passengers in a bus and the number of faces detects by the framework. We took different images at different, i.e., normal and complex
scenarios. Table 2 shows the actual number of passengers and detected faces by FDF.

To confirm the correctness and functioning of face detection, we compared the existing techniques such as SVM and HOG [11, 12] with the proposed solution. Table 2 shows that our results are more accurate in terms of face detection. In comparison with previous researches, the detection accuracy and performance is highly improved, i.e., 85% to 91%.

On the basis of these results, we can answer RQ1; Face detection based technique improves the existing APC with good accuracy. To answer RQ2 we compare the proposed and existing solution to counter the APC. [11, 12] calculate passengers using sensors. It is found to be expensive in terms of cost and high maintenance (REF). FDF is cost-effective because of the use of face detection technique and it requires low maintenance efforts (REF). As no previous research found from passenger perspective to solve the seat availability and reduce passenger time for waiting for the bus. The benefit gain from this research is to solve many problems, especially passenger behaviors in bus and traffic congestion problem.

Table 2. The detection rate of passengers faces

| Number of Passengers | Actual Faces | Detected Faces | Miss Faces | Percentage % |
|---|---|---|---|---|
| 27 | 25 | 24 | 1 | 96 |
| 32 | 32 | 30 | 2 | 93 |
| 38 | 36 | 34 | 2 | 94 |
| 43 | 43 | 41 | 2 | 95 |
| 49 | 49 | 45 | 4 | 91 |
| 55 | 53 | 48 | 5 | 90 |

## VI. CONCLUSION AND FUTURE WORK

The study proposes a passenger seat availability framework for integrating and processing passenger's faces, and then count passengers in a bus. It reduces traffic congestion problems, security risks, saving passenger time, passenger management and tracking. Our proposed framework is successfully simulated and tested in a real scenario. The results show that our system significantly reduces above problems. Moreover, it is cost effective and gives a good performance. It can be easily configured in bused and used by government authorities.

There are four limitations, which lead to opportunities for further work. First, we used an image dataset of passengers. Here, we didn't capture a live view of passenger and movements of buses. Thus, future research includes is to capture the live behavior of passengers. Second, firebase web server was used to store the information on routes, latitude, and longitude and bus stations. However, we used this trait for small work. Further research is to examine traits on own back-end server. Third, in face detection framework, we give an image of the passenger which takes two to three seconds every time. The whole process wasted two hours daily. Furthermore, this research will include automating the process which done automatically. Four, an app deploys locally for only iPhone users. Further research includes is to live in an app. We will develop an app for both Android and iOS devices.

**Khawar Islam** is undergraduate student.

**Uzma Afzal** is an Assistant Professor.